\documentclass{elsarticle}
\usepackage{hyperref}
\usepackage{graphicx}
\usepackage{subfigure}
\usepackage[cmex10]{amsmath}
\usepackage{amsfonts}
\usepackage{amssymb}

\makeatletter
\def\ps@pprintTitle{%
 \let\@oddhead\@empty
 \let\@evenhead\@empty
 \def\@oddfoot{}%
 \let\@evenfoot\@oddfoot}
\makeatother

\begin{document}

\begin{frontmatter}

\title{Holistic Interstitial Lung Disease Detection using Deep Convolutional Neural Networks: Multi-label Learning and Unordered Pooling}


\author{Mingchen Gao, Ziyue Xu, Le Lu, Adam P. Harrison, \\
Ronald M. Summers, Daniel J. Mollura\fnref{myfootnote}}
\fntext[myfootnote]{Mingchen Gao, Ziyue Xu, Adam P. Harrison, and Daniel J. Mollura are with Center for Infectious Disease Imaging; Le Lu and Ronald M. Summers are with the Imaging Biomarkers and Computer-Aided Diagnosis Laboratory and Clinical Image Processing Service. All authors are with the Radiology and Imaging Sciences Department, National Institutes of Health Clinical Center, Bethesda, MD 20892-1182, USA. }

%
%

\begin{abstract}
Accurately predicting and detecting interstitial lung disease (ILD) patterns given any computed tomography (CT) slice without any pre-processing prerequisites, such as manually delineated regions of interest (ROIs), is a clinically desirable, yet challenging goal. The majority of existing work relies on manually-provided ILD ROIs to extract sampled 2D image patches from CT slices and, from there, performs patch-based ILD categorization. Acquiring manual ROIs is labor intensive and serves as a bottleneck towards fully-automated CT imaging ILD screening over large-scale populations. Furthermore, despite the considerable high frequency of more than one ILD pattern on a single CT slice, previous works are only designed to detect one ILD pattern per slice or patch.

To tackle these two critical challenges, we present multi-label deep convolutional neural networks (CNNs) for detecting ILDs from holistic CT slices (instead of ROIs or sub-images). Conventional single-labeled CNN models can be augmented to cope with the possible presence of multiple ILD pattern labels, via 1) continuous-valued deep regression based robust norm loss functions or 2) a categorical objective as the sum of element-wise binary logistic losses. Our methods are evaluated and validated using a publicly available database of 658 patient CT scans under five-fold cross-validation, achieving promising performance on detecting four major ILD patterns: Ground Glass, Reticular, Honeycomb, and Emphysema. We also investigate the effectiveness of a CNN activation-based deep-feature encoding scheme using Fisher vector encoding, which treats ILD detection as spatially-unordered deep texture classification.

\end{abstract}


\end{frontmatter}


\section{Introduction}
\begin{figure*}[t]
\centering
\subfigure{\includegraphics[width=0.88\textwidth]{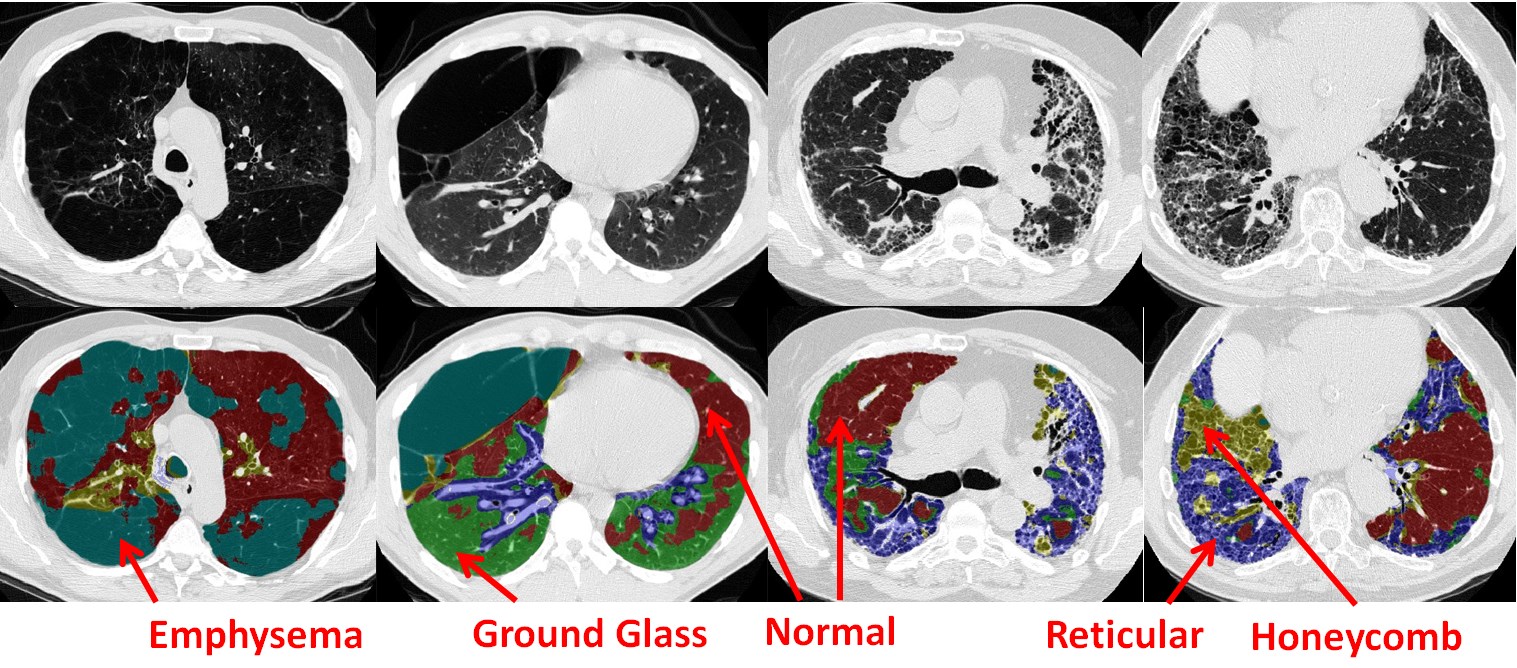}}
\caption{Examples of ILD patterns. Every voxel in the lung region is labeled as healthy or one of the four ILD patterns: Ground Glass, Reticular, Honeycomb, or Emphysema. The first row is the lung CT images. The second row refers to their corresponding labeling.}
\label{fig:examples}
\end{figure*}

Interstitial lung disease (ILD) refers to a group of more than 150 chronic lung diseases characterized by progressive scarring or inflammation of lung tissues and eventual impairment of breathing. The gold standard imaging modality for ILD diagnosis is computed tomography (CT)~\cite{depeursinge2012building,holmes2006lung}. Figure~\ref{fig:examples} depicts several examples of some most typical ILD-related CT imaging visual patterns. Automated detection of ILD patterns from CT images would aid the diagnosis and treatment of this morbidity.


The majority of previous work on ILD pattern detection is on 2D image classification at the patch level, which attempts to classify relatively small image patches (e.g., 32$\times$32 pixels) into one of the ILD pattern classes. These image patches are extracted or sampled from manually annotated polygon-like regions of interest (ROIs) on 2D axial slices, following the protocol of~\cite{depeursinge2012building}. Recent notable approaches include restricted Boltzmann machines~\cite{van2016combining}, convolutional neural networks (CNNs)~\cite{gao2015holistic,anthimopoulos2016lung,shin2016deep}, local binary patterns~\cite{song2015large,song2013feature} and multiple instance learning~\cite{hofmanninger2015mapping}. One prominent exception to the predominant patch-based approach is Gao \emph{et al.}'s work~\cite{gao2015holistic} which assigns a \emph{single} ILD class label directly upon whole axial CT slices, without any pre-processing to obtain ROIs.

When analyzing the Lung Tissue Research Consortium (LTRC) dataset~\cite{holmes2006lung}, which is the most comprehensive lung disease image database with per-pixel annotated segmentation masks, a significant number of CT slices are observed as being associated with two or more ILD labels. Despite the importance of predicting multiple possible ILD pattern types given an input CT image, this challenge has not been addressed by previous studies~\cite{anthimopoulos2016lung,van2016combining,gao2015holistic,song2015large,song2013feature}. ILD pattern detection is usually treated as a single-label classification problem from image patches~\cite{anthimopoulos2016lung,van2016combining,song2015large,song2013feature} or slices~\cite{gao2015holistic}.

Detecting multiple possible ILD types on holistic CT slices simultaneously arguably causes more technical challenges, but it results in a fully automated and clinically more realistic ILD classification process, especially when considering the problem of pre-screening large populations. Without knowing the actual ILD locations and regions of spatial extents \emph{a priori} (even lung segmentation), the methodological difficulties stem from several aspects, including 1) the tremendous amount of variation in ILD appearance, location and configuration; 2) the expense to obtain delicate pixel-level ILD annotations of large datasets for training and evaluation; and 3) the common occurrence of multiple ILD diseases coexisting on single CT slices. In this study, we target solving these three challenges at the same time.

One way to tackle the multi-label ILD recognition challenge is by replacing the softmax-based single-label loss~\cite{krizhevsky2012imagenet} with a multi-label classification loss layer (Sec. \ref{sec:mulc}). Our method works on a holistic CT slice as an input to directly provide multiple ILD patterns existing on that slice, which moves forward an important step to delivering better clinical relevance than previous work~\cite{gao2016Multi,gao2016Segmentation}. Alternatively, partially inspired by the recent natural image classification work~\cite{wei2014cnn}, we explore an alternative method, which models this multi-label prediction problem using a continuously valued regression formulation (Sec. \ref{sec:mulr}). Note that multi-label regression has also been used outside of ILD contexts to estimate heart chamber volume~\cite{Zhen2015Direct,Zhen2014Direct}.

We employ the end-to-end deep CNN regression model because of its simplicity and the fact that deep image features and final cost functions can be learned simultaneously~\cite{krizhevsky2012imagenet,vedaldi2015matconvnet}. End-to-end deep neural network representations have shown significant performance superiority over the variants of ``hand-crafted image features followed by a separate classifier'', in recent studies~\cite{Bar2016Chest,Ginneken2016Off}.

While CNNs are powerful image recognition models, its deep image feature learning and encoding representation is not invariant to the spatial locations and layouts of objects or texture patterns within a holistic visual scene (e.g., an input CT slice). As observed in~\cite{cimpoi2016deep,gong2014multi}, this order-sensitive CNN feature encoding, reflecting the spatial layout of the local image descriptors, is effective in object and scene recognition but may not be beneficial, or can even be counter-productive, for texture classification. The default order-sensitive spatial encoding of CNN image descriptors can be removed through the schemes of unordered feature encoders, such as bag of visual words (BoVW), Fisher vectors (FV)~\cite{perronnin2010improving}, or aggregation of spatial pyramid matching (SPM)~\cite{gong2014multi}, etc. Previous work on image patch based approaches~\cite{anthimopoulos2016lung,van2016combining,song2015large,song2013feature}, are equivalent to formulating ILD pattern recognition as texture classification since the gross image layout information is discarded. Therefore, given the above considerations, we attempt to answer the question whether ILD recognition is indeed a texture classification problem by performing spatially invariant feature encoding from image feature activations from the CNN regression architecture, followed by dimension reduction and multivariate linear regression (Sec. \ref{sec:up}).

Our methods are validated using the publicly available LTRC ILD dataset~\cite{holmes2006lung}, composed of 658 patients which are all the data in LTRC consisting of good ILD annotations. Our experiment protocol employs five-fold cross-validation (CV) to detect the most common ILD classes of Ground Glass, Reticular, Honeycomb, and Emphysema. Extensive quantitative experimental results show the promise of our approach in tackling the challenges of multi-class ILD classification given any input CT slice, without any manual pre-processing.


\section{Related work}
Detecting ILD patterns in CT imaging is commonly treated as a texture recognition and classification problem in many previous studies \cite{sluimer2003computer,song2013feature,anthimopoulos2014classification,bagesteiro2015blockwise,sorensen2010quantitative}. Moreover, texture based visual representation is adopted inside local image regions of interest (ROIs) or volumes of interest (VOIs) via extracting rectangular image patches, when a 2D and 3D CT imaging modality is used, respectively. In the sliding window manner, image classifiers can generate an ILD probability map within a pre-segmented lung region. Image feature extraction and machine learning based classification are two separate factors in building previous image recognition systems.

An early work on computer-aided ILD recognition is proposed to employ neural networks and expert rules to detect ground glass opacity (GGO) on CT images~\cite{heitmann1997automatic}. The follow-up work includes GGO detection and segmentation~\cite{zhou2006automatic,Tao2009Multi}. Shyu et al.~\cite{shyu1999assert} describe a human-in-the-loop approach where the human annotator delineates the region of interest and anatomical landmarks in the images, followed by classification on image attributes related to variations in intensity, texture, shape descriptors and so on. Zheng et al.~\cite{zheng2004automated} analyze 3D ILD imaging regions that are combined from multiple candidates detected beforehand on 2D slices. Fukushima et al.~\cite{fukushima2004application} evaluate the diagnostic performance of an artificial neural network.

There are many types of hand-crafted image features that are adopted for ILD classification, such as filter banks\cite{sluimer2003computer,song2013feature,anthimopoulos2014classification}, local binary patterns (LBPs)~\cite{bagesteiro2015blockwise,sorensen2010quantitative}, morphological operators followed by geometric measures, histogram of oriented gradients~\cite{song2013feature}, texton based approaches~\cite{gangeh2010texton}, and wavelet and contourlet transforms~\cite{vo2010multiple,depeursinge2012near}. 2D texture features have also been extended into three dimensions~\cite{xu2006computer, depeursinge2015optimized,Tao2009Multi}. Typical feature encoding scheme and classifiers include bag of words~\cite{xu2011classification}, support vector machines (SVMs)~\cite{bagesteiro2015blockwise, xu2006computer, depeursinge2012near}, random forest~\cite{anthimopoulos2014classification} and k-nearest neighbors (kNN)~\cite{sluimer2003computer}.

In contrast to separate hand-crafted features and classifier modeling, convolutional neural networks (CNN) can learn image features and the classifier simultaneously. Restricted Boltzmann machines (RBMs) have been used to learn unsupervised classification features within lung regions~\cite{li2013lung}, whereas CNNs are used in a supervised formulation ~\cite{li2014medical}. In~\cite{van2016combining}, a convolutional classification RBM is trained combining a generative and a discriminative learning objective. \cite{anthimopoulos2016lung} proposes a specially designed CNN architecture for the classification of ILD patterns. This network consists of five convolutional layers with $2 \times 2$ kernels and LeakyReLU activations, followed by average pooling and three fully connected layers. The size of the kernels in each layer is chosen to be minimal, which leads to deep networks, similar to VGG-net~\cite{simonyan2014very}. \cite{shin2016deep} articulates several important approaches toward employing CNNs in medical imaging applications. The ILD pattern classification problem was explored and evaluated using different CNN architectures. In particular, transfer learning was studied using pre-trained ImageNet CNN models \cite{krizhevsky2012imagenet} to fine-tune on domain-specific tasks of medical imaging detection and diagnosis.

A preliminary version of this work appears in \cite{gao2016Multi}. In this paper, we propose, extend and fully evaluate two different multi-label CNN classification architectures to address the phenomenon of multiple ILDs' co-occurrence on single CT images. Robust deep regression loss function under multi-label setting is also addressed. The improved algorithms are extensively validated with a more complete dataset, using comprehensive evaluation metrics, and by conducting comparable experiments against patch based ILD classification, which constitutes the majority of previous work. Superior quantitative performance in both detection accuracy and time efficiency is demonstrated.

\section{Methods}
In this section, we propose three variations of multi-label deep convolutional neural network classification or regression models to address the multi-label ILD detection challenge. First, an end-to-end CNN network is trained using a multi-label image classification loss layer. Second, we outline a CNN network that uses a continuously-valued regression formulation, estimating either the actual pixel numbers occupied per ILD class per CT image or the binary  [0,1] occurring status. Third, the convolutional image activation feature maps at different network depths are spatially aggregated and encoded through the orderless Fisher vector (FV) encoder~\cite{perronnin2010improving}. This encoding scheme removes the spatial configurations/layouts of convolutional activations and turns them into location-invariant feature representations. This type of CNN is referred to as FV-CNN~\cite{cimpoi2016deep}. The formed orderless features are then trained with a multivariate linear regressor (Mvregress($\ast$) function in Matlab) to regress the ILD pixel numbers or binary labels.

\begin{figure*}[hbtp]
\centering
\subfigure{\includegraphics[width=1.05\textwidth]{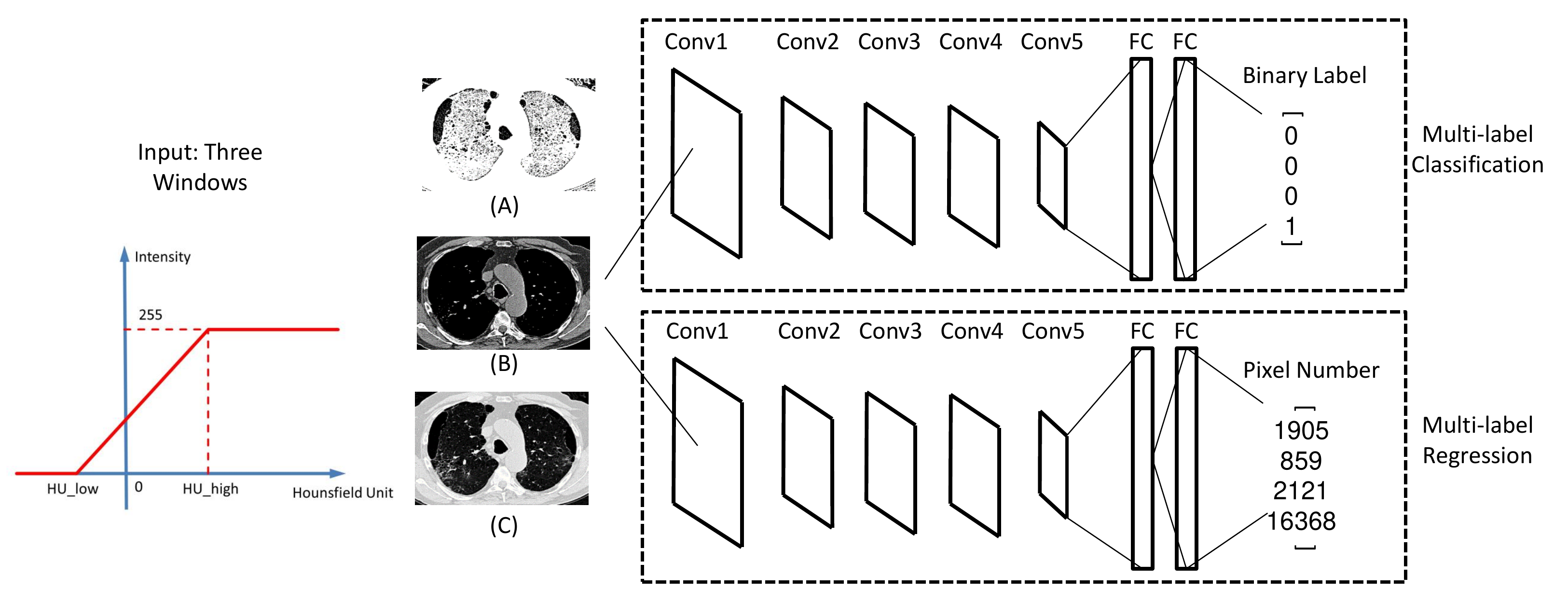}}
\caption{Multi-label CNN models: the input CT slices are transformed into three attenuation scales from the original CT value range that can highlight different anatomical tissues and lung disease patterns), before being fed into CNN for processing. Three CNN models are proposed to detect possibly multiple ILD diseases per CT image ($512\times512$ pixel CT slice instead of smaller image ROIs or patches in most previous work). The soft-max layer of single-label multi-class CNN is replaced by either a multi-label classification CNN loss layer (Sec. \ref{sec:mulc}) or real-valued regression loss layer (Sec. \ref{sec:mulr}).}
\label{fig:flowchart}
\end{figure*}

There are several mainstream CNN architectures, such as AlexNet~\cite{krizhevsky2012imagenet}, VGGNet~\cite{simonyan2014very}, GoogLeNet~\cite{szegedy2015going}, and deep residual networks~\cite{he2015deep}. Each network has its own advantages and is suitable for specific applications. Here we employ a variation of AlexNet, called CNN-F~\cite{chatfield2014return}, for its good trade-off between efficiency and performance. Fully-annotated medical imaging datasets are usually of limited availability and can be much smaller than the popular computer vision ImageNet database~\cite{russakovsky2015imagenet}. The classical CNN-F contains five convolutional layers, followed by two fully-connected (FC) layers, and a last softmax layer for classification. We modify it to accommodate our application of detecting multiple ILD patterns in CT images, as shown in Fig.~\ref{fig:flowchart}. Based on our empirical test using a much deeper CNN model of VGG-19, deeper models do not provide significantly noticeable quantitative performance boosts in ILD classification accuracy while at the same time they consume much more training and testing time.

Our three main deep learning algorithms, namely multi-label CNN classification, robust deep regression, and unordered pooling multivariate regression, are described in Sec.~\ref{sec:mulc}, Sec.~\ref{sec:mulr} and Sec.~\ref{sec:up}, respectively. Two additional critical technical aspects are then addressed, i.e., balancing the distribution of different classes to achieve the performance boost (Sec.~\ref{sec:class_balancing}), and exploiting different CT attenuation scaling schemes to better capture the visual appearance of abnormal ILD patterns (Sec.~\ref{sec:rescale}).


\subsection{Multi-label Classification Deep CNN}\label{sec:mulc}

The multi-label classification problem has not been widely studied in ILD recognition because in previous image patch based representation, each patch belongs to a single ILD class making it unnecessary to add the additional complexity of multi-labeling. In the holistic CT slice based ILD prediction, we first tackle this task using a multi-label classification CNN loss, under a one-against-all formulation (Eq. \ref{eq:mlcl}), as studied in the recent computer vision literature~\cite{oquab2015object,durand2016weldon,li2015weakly}. The typical logistic regression loss in the soft-max CNN layer is only capable of predicting a single label for each instance, which is not suitable to our application. Our goal is to identify the existence of all occurring ILDs simultaneously where the input image may contain multiple ILD patterns. Our employed loss function is intuitive and effective in the form of a sum of $C$ binary logistic regression losses, one for each of the $C$ classes $k \in \{1 \ldots C\}$,
\begin{equation}\label{eq:mlcl}
L(y, f(x)) = \sum_{k=1}^C log(1+exp(-y_kf_k(x))),
\end{equation}
where $y_k \in \{-1,1\}$ is the label indicating the absence/presence of class $k$ given input image $x$ and $f_k(x)$ is the output of the network.


In each loss computation per input image, the logistic loss and gradients according to all ``positive'' labels (on that image) are calculated and added within the network loss layer intrinsically, for the stochastic gradient back-propagation (BP) of neural network training. There is another possible design choice, which is to treat the multi-label classification problem as $C$ independent multi-task classification problems. In contrast to Eq. \ref{eq:mlcl}, we need to have $C$ separate binary loss layers to cover all ILD classes and the gradient BP process during training will be independently computed. Thus this ``multi-head'' multi-mask CNN can not model the intrinsic correlations among multiple ILD labels, which based on our empirical finding, causes the multi-mask network hard to converge.

\subsection{Robust Multi-label Regression Deep CNN}\label{sec:mulr}

Next, we propose and investigate this multi-label problem via deep regression losses. Supposing that there are in total $N$ images and $C$ types of ILDs to be detected, the label vector for an image $I$ is represented as a $C$-dimensional label vector $\boldsymbol{y} = [y_{1}, \ldots, y_{k}, \ldots, y_{C}] \in \{0,1\}^C, k \in \{1,\ldots,C\}$, in which each entry can be $1$ or $0$, indicating whether a specific disease exists in the image. For the case of one image containing multiple diseases, there will several corresponding $1$'s in the label vector $\boldsymbol{y}$. An all-zero $\boldsymbol{y}$ represents a healthy slice/instance, \emph{\i.e.}, no targeted ILD being found. Alternately, the actual number of pixels of each ILD pattern per image can be recorded in $\boldsymbol{y}$, replacing the binary value of $1$ with an integer quantity.
This multivariate label vector allows our algorithm to naturally preserve the frequent co-occurrence property of ILDs in CT imaging through (deep) regression. We summarize the different options below. 

{\bf Loss Functions:} Deep CNN regression loss is used to learn the presence or the spatial occupancy area (in terms of the pixel number) of ILD classes per image. The following loss function (Eq. \ref{eq:regression})~\cite{wei2014cnn,girshick2015fast} is adopted instead of the more widely used softmax loss for classification CNNs~\cite{gao2015holistic,anthimopoulos2016lung,krizhevsky2012imagenet}. The loss cost function to be minimized is defined as
\begin{equation}\label{eq:regression}
L({y}, f(x)) = \sum_{k=1}^C \text{L}_\text{reg}(y_{k}-f(x)),
\end{equation}
where $\text{L}_\text{reg}$ could be either $L_2 = x^2$ loss or smooth $L_1$ loss.  The smooth $L_1$ cost function~\cite{girshick2015fast} is defined as
\begin{equation}
\text{smooth}_{L1}(x) = \left\{
\begin{array}{rl}
0.5 x^2 & \text{if } |x| < 1,\\
|x|-0.5 & \text{otherwise}.
\end{array}
\right.
\label{eq:smoothl1}
\end{equation}
This robust smooth $L_1$ loss is designed to be less numerically sensitive to outliers (extremely large targeted label values) than $L_2$ loss. The use of Eq.~\ref{eq:smoothl1} could eliminate the chances of exploding gradients, to which the $L_2$ loss is subject.

\begin{figure*}[hbtp]
\centering
\subfigure{\includegraphics[width=1\textwidth]{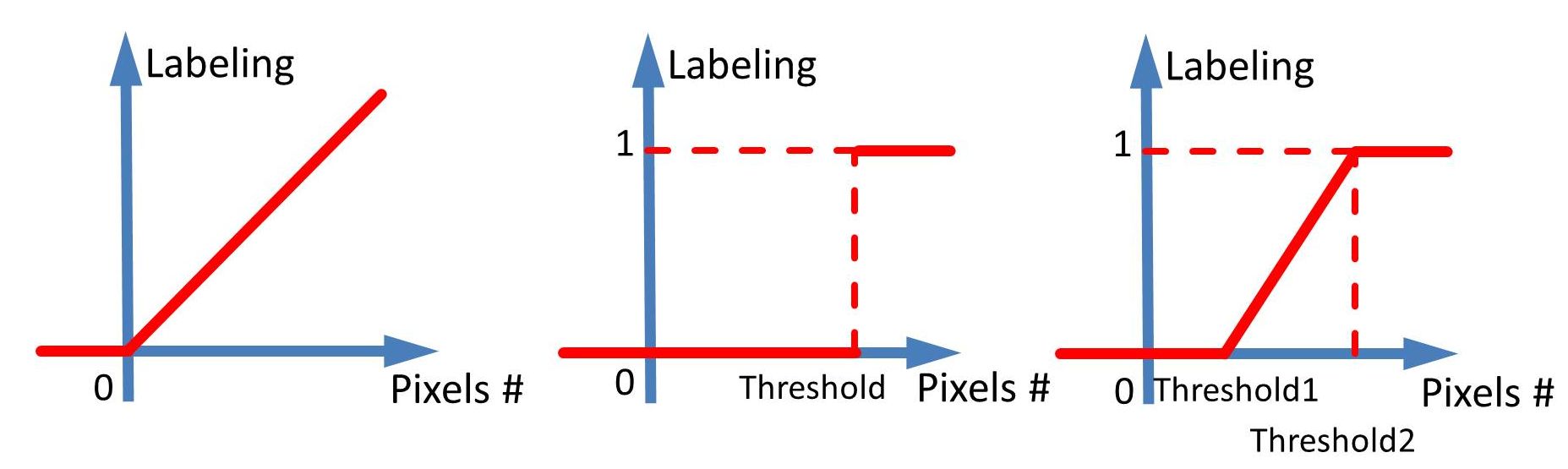}}
\caption{Three mapping functions transfer the pixel counts (per ILD class) to their label values for training the CNN regression losses.}
\label{fig:labeling_function}
\end{figure*}

{\bf Binary or Continuously Valued Regression:}
There are several options to form the regression labels for each image. One straightforward scheme is to count the total number of pixels annotated per ILD disease, which represents its severity (Fig.~\ref{fig:labeling_function} {\emph Left}). The step function to represent the presence (1) or absence (0) of the disease (Fig.~\ref{fig:labeling_function} {\emph Middle}) is also possible. The binarizing threshold $T$ may be defined using clinical knowledge: if the pixel number is $\geq T$, the label is set to be 1; otherwise as 0. A more sophisticated label transfer model is a piecewise linear function of the pixel counts with $T_1, T_2$, mapping pixel counts to the a range of $[0,1]$ (Fig.~\ref{fig:labeling_function} {\emph Right}). Diseases with the number of total pixels $\geq T_1$ but $\leq T_2$ are linearly interpolated to between $0$ and $1$.

\subsection{Unordered Pooling via Fisher Vector Encoding for Multivariate Linear Regression}\label{sec:up}

\begin{figure}[hbtp]
\centering
\subfigure{\includegraphics[width=1.1\textwidth]{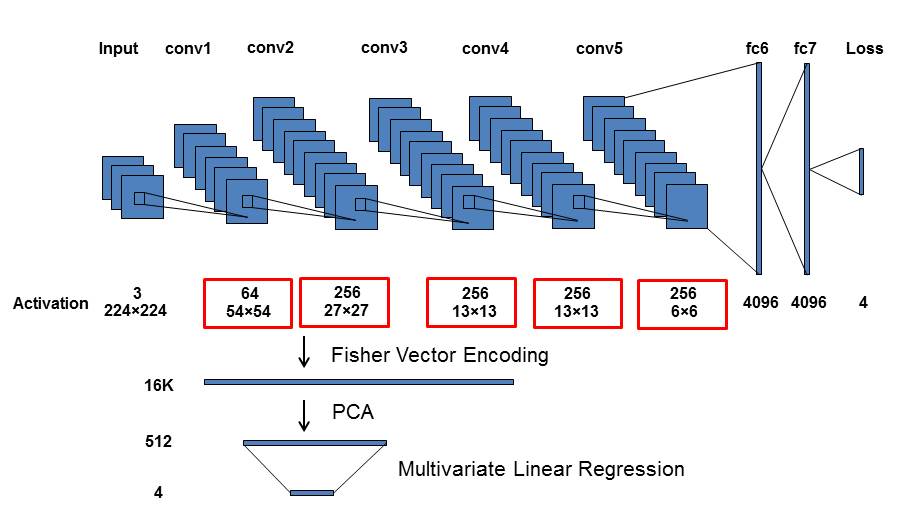}}
\caption{The overall architecture of the CNN model and unordered polling using FV encoding and multivariate linear regression. }
\label{fig:FV_example}
\end{figure}

Both classification and regression CNN models (Sec. \ref{sec:mulc} and \ref{sec:mulr}) can be seen as performing the spatially order-sensitive feature pooling through their use of fully-connected CNN layers. In this section, we investigate whether the spatial information captured inside CNN activation maps is beneficial for a task-specific image recognition problem. The typical representation of deep hierarchical CNN features inherits the gross image-activation spatial layouts. We are motivated by the observation that ILD patterns could happen anywhere inside the lung region, implying that the spatial layout may not be a strongly-correlated factor to ILD recognition. CNNs are designed to learn special feature layouts from the limited annotated ILD imaging data, which may be subject to over-fitting more easily. 
In our implementation, CNN activations are extracted from the convolutional layers at various depths of the CNN network and compiled using the Fisher vector (FV) feature encoding scheme \cite{perronnin2010improving,cimpoi2016deep}, allowing us to achieve a location-invariant deep texture description. ILD class labels can then be predicted via the simple multivariate linear regression. 


The output of each $k$-th convolutional layer is a 3D descriptor or matrix $\boldsymbol{X_k} \in \mathbb{R}^{W_k\times H_k\times D_k} $, where $W_k$ and $H_k$ are the width and height of the spatial reception field and $D_k$ is the number of feature channels. In this sense, the specific deep feature activation map is represented by $W_k \times H_k$ feature vectors and each feature vector is $D_k$ dimension. We invoke the FV encoding to remove the spatial configurations of total $W_k \times H_k$ vectors (denoted as the set $\boldsymbol{X_k}$) for each activation map. Following \cite{perronnin2010improving}, each feature descriptor $x_i \in \boldsymbol{X_k}$ is soft-quantized using a Gaussian mixture model. The first- and second-order differences $(u_{i,m}^T, v_{i,m}^T)$ between any descriptor $x_i$ and each of the Gaussian cluster mean vectors $\{ \boldsymbol{\mu _m} \}, m=1,2,...,M$ are accumulated into a $2MD_k$-dimensional image representation:
\begin{align}
\boldsymbol{f^{FV}_i} = [u_{i,1}^T, v_{i,1}^T,...,u_{i,M}^T, v_{i,M}^T] \textrm{.}
\end{align}

FV feature encoding produces a very high dimensionality of $2MD_k$ from the deep feature activations of $\boldsymbol{X_k}$ for each image (e.g., $M=32$ and $D_k=256$). For computational and memory efficiency, we adopt principal component analysis (PCA) to project the $\boldsymbol{f^{FV}_i}$ vectors to a lower-dimensional parameter subspace. Using their ground-truth ILD label vectors $\boldsymbol{y_i}$, the multivariate linear regression Mvregress($\ast$) function in Matlab is called to predict the presence or non-presence of ILDs from the low-dimensional projected image feature vectors $PCA(\boldsymbol{f^{FV}_i})$.

\subsection{Class Balancing}\label{sec:class_balancing}

Most computer-aided detection datasets have very biased distributions for instances of different classes. In our setting, some ILD types may appear more often than others. We utilize a simple and effective strategy to automatically balance the loss among different classes~\cite{xie2015holistically}. A class-balancing weight $\beta_k$ for each class $k$ is added into the classification CNN loss.
\begin{equation}
\label{eq:classification_balancing}
L(y, f(x)) = \sum_{k=1}^C \beta_k log(1+exp(-y_kf_k(x))),
\end{equation}
Similarly, the weighted multi-label regression loss is
\begin{equation} \label{eq:regression_balancing}
L({y}, f(x)) = \sum_{k=1}^C \beta_k \text{L}_\text{reg}(y_{k}-f(x)),
\end{equation}
where $\beta_k = \frac{1-|Y_k|/|Y|}{C}$, $|Y_k|$ denotes the cardinality of the dataset of class $k$ according to the ground truth labels, and $|Y|=\sum_{k=1}^C |Y_k|$.

\subsection{CT Attenuation Rescaling}\label{sec:rescale}

To better capture the abnormal ILD patterns in CT images, we select three CT attenuation ranges or windows and rescale them to $[0, 255]$ for CNN input. This is inspired by the fact that radiologists adjust the CT contrast window to optimize the visualization effects for certain tissues or pathologies on CT scans. For example, ``Lung'' window visualizes details in lung tissue that are not apparent on the ``Bone'' window, whereas much of the information in bone and soft tissue will be lost on the ``Lung'' window. In this work, we use three CT attenuation scales to highlight different lung disease patterns. As demonstrated in Fig.~\ref{fig:flowchart}(A), this process is designated to preserve the attenuation values between HU\_low and HU\_high via a linear transformation. The intensity values outside the specific attenuation window are set as 0 or 255. In details, the low attenuation range (Fig.~\ref{fig:flowchart}(B)) is used to capture ILD patterns with lower intensities, such as emphysema; the normal range (Fig.~\ref{fig:flowchart}(C)) to represent normal appearance of lung regions; and high attenuation range for highlighting patterns with higher intensities, for example, consolidation and nodules. In our experiments, the low attenuation window is set as HU\_low$=-1400$ and HU\_high$=-950$; for normal range, HU\_low$=-1400$ and HU\_high$=200$; for high attenuation scale, HU\_low$=-160$ and HU\_high$=240$.

\section{Experiments and Discussion}

\subsection{Data}
There are two main publicly available datasets for CT imaging based ILD classification~\cite{depeursinge2012building,holmes2006lung}. The LTRC~\cite{holmes2006lung} dataset provides complete ILD labeling at the per-voxel or per-pixel level. In contrast, missing labels are common in \cite{depeursinge2012building}. Additionally, not all ILD regions of interest per-slice are delineated by annotators and often only one prominent disease region is annotated on a slice, as studied in ~\cite{gao2016Segmentation,depeursinge2012building}. As a result, we use the LTRC dataset for our method validation and performance evaluation. Every voxel inside the CT lung region is labeled as healthy or one of the four ILD types: Ground Glass, Reticular, Honeycomb or Emphysema. Our goal is, given any input axial CT slice, to predict ILD labels where multiple diseases could co-occur. The number of classes $C$ is set as 4 to represent the four ILD types.

For ease of comparison, 2D axial CT slices or images are evaluated independently, without taking successive slices into consideration. Many CT scans of ILD study can have large inter-slice distances, for example, 10mm in~\cite{depeursinge2012building} between successive axial slices, making direct 3D volumetric analysis implausible. Utilizing only 2D axial image information makes the algorithm more generalizable to low-dose CT imaging based ILD screening protocols.

In total, there are 658 patients in the LTRC dataset for ILD classification and detection. Some ILD patients, which cannot find matched annotations with CT images, are eliminated. The original resolution of the 2D axial slices is $512 \times 512$ pixels. All images are resized to the uniform size of $224 \times 224$ pixels. Five-fold cross-validation (split at patient level) is conducted for the quantitative experimental evaluation.
There are $\sim$240k CT slices in total from the 658 patients for cross-validation (CV). CNN training is performed in Matlab using MatConvNet~\cite{vedaldi2015matconvnet} and run on a PC  with 3.1GHz CPU, 32 GB memory, and an Nvidia Tesla K40 GPU.

\subsection{Multi-label Classification Evaluation}
\label{section:evaluations}
Single label ILD classification can be quantitatively evaluated using recall, precision, and F-score metrics, respectively for each disease. Multi-label classification needs different performance metrics than those used in the single-label scenario~\cite{tsoumakas2006multi}. Let $T$ denote the ground truth set of labels; and $S$ be the predicted set. Accuracy is measured by the Hamming score which is symmetrical measurement of how close $T$ is to $S$, as illustrated in  Eq.~\ref{eq:multilabel_metrics_accuracy}. Similar formulations are applied to calculate precision and recall under multi-label classification evaluation (Eq. \ref{eq:multilabel_metrics_precision} and \ref{eq:multilabel_metrics_recall}). F-score, which is the harmonic mean of precision and recall, keeps the same for both single and multi-label classification evaluations (Eq. \ref{eq:multilabel_metrics_F}). In our experiments, we evaluate the overall multi-label ILD prediction performance and report the results for each individual ILD as well.


\begin{equation}
    \text{Accuracy}(T,S) = \frac{1}{n} \sum_{i = 1}^n \frac{|T_i \cap S_i|}{|T_i \cup S_i|},\label{eq:multilabel_metrics_accuracy}
\end{equation}
\begin{equation}
    \text{Precision}(T,S) = \frac{1}{n} \sum_{i = 1}^n \frac{|T_i \cap S_i|}{|S_i|}, \label{eq:multilabel_metrics_precision}
\end{equation}
\begin{equation}
    \text{Recall}(T,S) = \frac{1}{n} \sum_{i = 1}^n \frac{|T_i \cap S_i|}{|T_i|},
    \label{eq:multilabel_metrics_recall}
\end{equation}
\begin{equation}
    F_1 = 2 \times \frac{Precision \times Recall}{Precision + Recall},
    \label{eq:multilabel_metrics_F}
\end{equation}

\subsection{Results on Multi-label Classification CNN} \label{exp:classification}

To conduct holistic CT slice based ILD classification, we first convert the pixel-wise annotated masks in LTRC~\cite{holmes2006lung} into slice-level labels. Without loss of generality, we set the pathology threshold $T=6000$ or $T=4000$ pixels to differentiate the presence (if $\geq T$) or absence (if $\leq T$) of ILDs. The number of slices containing each ILD pattern is outlined in Table~\ref{table:stats} when the pathology presence threshold is set as $T = 6000$. There are many CT slices or instances with multiple ILDs co-existing on the same slice, as shown in Table~\ref{table:multiple_diseases} .

\begin{table}
 \caption{Statistics on the 658 patients and 240k CT slices from LTRC~\cite{holmes2006lung}. Without loss of generality, threshold $T=6000$ pixels is used to differentiate the presence or absence of ILD patterns.} \label{table:stats}.
\centering
    \setlength{\tabcolsep}{5pt}
    \begin{tabular}{  c | c  c }
    \hline
    \textbf{ILD pattern}  & Positive   & Negative\\ \hline
    Healthy & 226675&        15362  \\ 
    Ground Glass & 41194     &200843   \\ 
    Reticular & 20560  &    221477  \\ 
    Honeycomb & 17392  &    224645 \\ 
    Emphysema & 36328  &    205709 \\ \hline
    \end{tabular}
\end{table}

\begin{table}
 \caption{The number of slices with multiple ILD patterns coexisting on the same slice. Pathology presence threshold $T=6000$ pixels is used.} \label{table:multiple_diseases}
\centering
    \setlength{\tabcolsep}{5pt}
    \begin{tabular}{  c | c | c | c | c}
    \hline
    {Healthy}  & One Disease & Two Diseases  & Three Diseases & Four Diseases\\ \hline
    149950 & 70339 & 20127 & 1603 & 18 \\ \hline
    \end{tabular}
\end{table}

\begin{figure*}[!ht]
\centering
\subfigure[]{\includegraphics[width=0.49\textwidth]{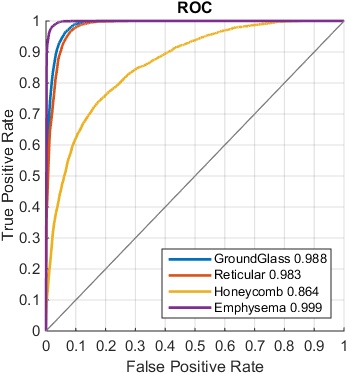}}
\subfigure[]{\includegraphics[width=0.49\textwidth]{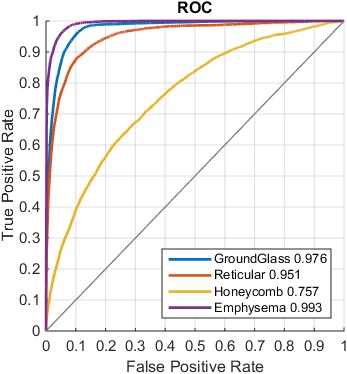}}\\
\subfigure[]{\includegraphics[width=0.49\textwidth]{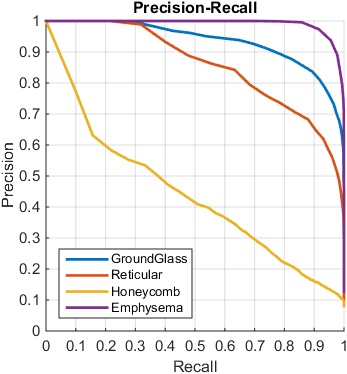}}
\subfigure[]{\includegraphics[width=0.49\textwidth]{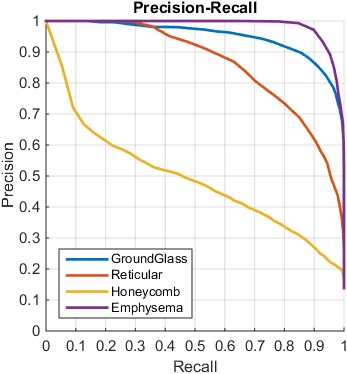}}\\
\caption{ILD classification results shown in ROC and Precision-Recall curves using the threshold to 6000 pixels in (a)(c) and 4000 pixels in (b)(d).}
\label{fig:ROC_classification_scores}
\end{figure*}

The classification results are shown using the ROC curves in Fig.~\ref{fig:ROC_classification_scores}, and F-scores are presented in Table~\ref{table:F-scores-6k} and Table~\ref{table:F-scores-4k} while setting the pathology presence thresholds to be $T=6000$ and $T=4000$, respectively. The overall F-score is calculated based on the multi-label classification evaluation mentioned in Sec. \ref{section:evaluations}. We obtain good results using the setting of $T=4000$ but the quantitative results by setting $T=6000$ are further improved, indicating that our algorithm may be robust to detect smaller ILD patterns and can tolerate some pixel-level annotation errors in LTRC. In our setting, pixel-level ILD annotations are not essentially required. Therefore the medical experts can simply provide the holistic CT slice-level labels on any lung CT image to indicate if there are ILD presences worth reporting, without annotating particular ILD image ROIs. It would considerably save the labeling time for experts to annotate the training dataset.

\begin{table}
\caption{F scores of multi-label classification and regression CNNs, with the setting $T=6000$ pixels.} \label{table:F-scores-6k}
\centering
    \setlength{\tabcolsep}{5pt}
    \begin{tabular}{   c | c  c  c  c  c }
        \hline
         & \multicolumn{5}{c}{\textbf{F-score}} \\ \hline
         \textbf{Disease}  & \textbf{Ground Glass}  & \textbf{Reticular} & \textbf{Honeycomb} & \textbf{Emphysema} & \textbf{Overall}\\ \hline
         classification  & {0.8642} & \textbf{0.7686} & \textbf{0.4602} & \textbf{0.9468}  & 0.7959  \\

         regression $L_2$ loss & {0.8343} & 0.4764 & 0.2314 & 0.8042 & 0.6882   \\ 
         regression smooth $L_1$ loss & \textbf{0.9102} & {0.7095} & {0.3385} & {0.8991}  & \textbf{0.8028}  \\
        \hline
    \end{tabular}
\end{table}

\begin{table*}
\caption{F scores of multi-label classification and regression CNNs, with the setting $T=4000$ pixels.} \label{table:F-scores-4k}
\centering
    \setlength{\tabcolsep}{5pt}
    \begin{tabular}{ c | c  c  c  c  c }
        \hline
        & \multicolumn{5}{c}{\textbf{F-score}} \\ \hline
        \textbf{Disease}  & \textbf{Ground Glass}  & \textbf{Reticular} & \textbf{Honeycomb} & \textbf{Emphysema} & \textbf{Overall}\\ \hline

        classification   & {0.8825} &\textbf{ 0.7667} & \textbf{0.5068} & \textbf{0.9361 } & {0.7656 }\\

        regression $L_2$ loss & {0.8385} & 0.5868 & 0.3102 & 0.8008 & 0.6487   \\
        regression smooth $L_1$ loss  & \textbf{0.9079} & 0.7190 & 0.4092 & {0.9152}  & \textbf{0.7774}  \\
        \hline


    \end{tabular}
\end{table*}

\subsection{Results on Multi-label Regression CNN}

We can treat the continuously-valued output vector, either in the form of pixel number counts or binary presence status, as the ``classification confidence scores'' after the multi-label regression CNN processes an input CT image. These regressed confidence scores can be compared against the ground truth binary ILD labels obtained by thresholding on $T$ as in Sec. \ref{exp:classification}. In this manner, ILD classification receiver operating characteristic (ROC) curves are generated. Our experiments are conducted via the three label converting functions or plots in Fig.~\ref{fig:labeling_function}. Two variations of CNN regression outputs to match the ILD occupied pixel numbers per-slice, or the binary ILD presence labels produce similar quantitative ILD classification results. The piecewise linear transformation (Fig. \ref{fig:labeling_function} {\emph Right}) yields slightly inferior results.

\begin{figure*}[t]
\centering
\subfigure{\includegraphics[width=1.05\textwidth]{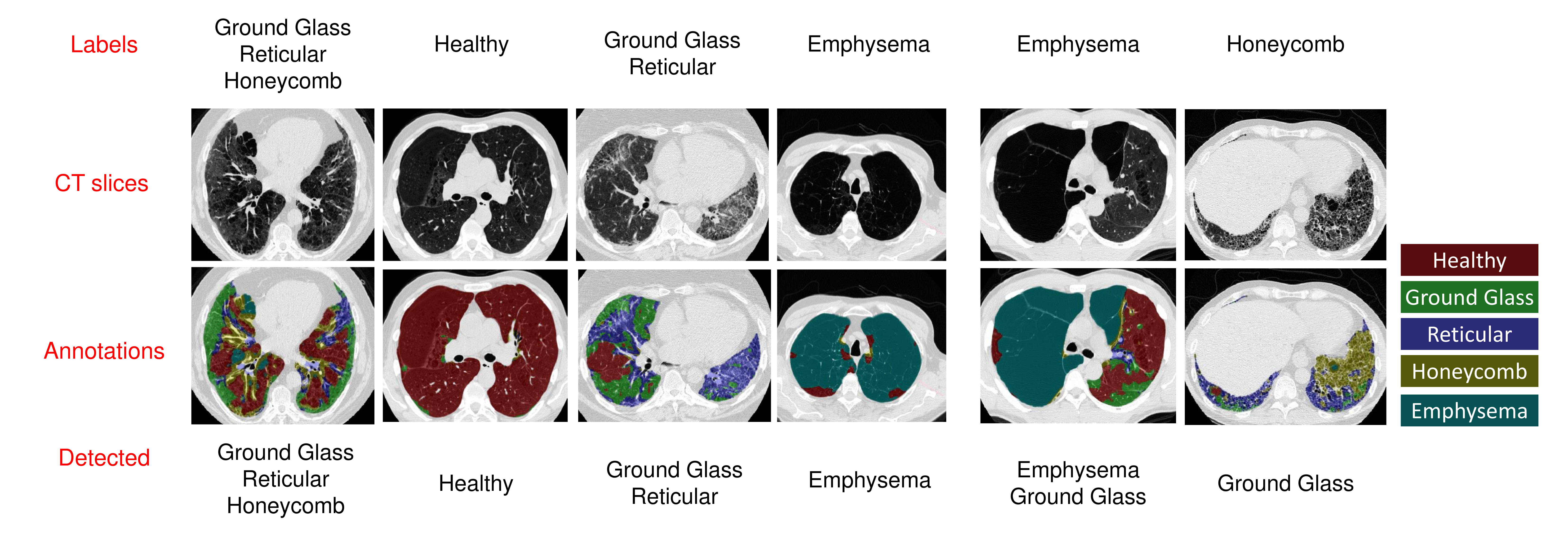}}
\caption{Examples of successfully detected and misclassified ILD slices. The left four are the correctly labeled cases, and the right two are failed cases.}
\label{fig:results}
\end{figure*}

Table~\ref{table:F-scores-6k} and Table~\ref{table:F-scores-4k} show the multi-label regression CNN results where the trained model regresses to the number of diseased pixels in each image. The use of a smooth $L_1$ cost function greatly improves the performance  and constantly outperforms the $L_2$ cost function in all experiments by noticeably large margins. Fig.~\ref{fig:results} illustrates some visual examples of successful or misclassified results. The first four examples are successfully detected cases, with multiple ILD patterns coexisting on the same slice. The last two are failure cases. Note that the first misclassified case is marked with two detected labels of ``emphysema'' and ``ground glass''. Both emphysema and ground glass co-occur on this image but the pixel count of ground glass occupied spatial region does not meet the pathology threshold of $\geq T=6000$. These qualitative results visually confirm the high performance demonstrated by our quantitative evaluation.

The overall performance of multi-label regression CNN when the smooth $L_1$ loss is employed is generally comparable with the multi-label classification CNN (Sec. \ref{exp:classification}). From Table \ref{table:F-scores-4k} and \ref{table:F-scores-6k}, the smooth $L_1$ regression CNN performs slightly better overall and particularly for the ground glass class, but the multi-label classification CNN outperforms in the categories of reticular, honeycomb, and emphysema with moderate margins.

\subsection{Unordered Pooling via Fisher Vector Encoding}

When constructing the FV-encoded features, $\boldsymbol{f^{FV}_i}$, the local convolutional image descriptors are pooled into $32$ Gaussian components, producing a dimensionality as high as 16K \cite{perronnin2010improving}. We further reduce the FV features to $512$ dimensions using PCA. The performance is empirically found to be insensitive to the number of Gaussian kernels and the dimensions after PCA. We compare the ILD classification performance with FV encoding, on the features pooled from different CNN layers, using area-under-the-curve (AUC) values (in Table~\ref{table:AUC_pixels_pixels}) and F-scores (in Table \ref{table:F_pixels_pixels}), respectively.

\begin{table}
 \caption{AUC values among different CNN layers by FV encoding and linear regression. Both CNN and multi-variant linear regression regress to ILD pixel numbers.} \label{table:AUC_pixels_pixels}
\centering
    \setlength{\tabcolsep}{5pt}
    \begin{tabular}{  c | c  c  c  c  c  c  c }
    \hline
    & \multicolumn{6}{c}{\textbf{AUC}} \\ \hline
    \textbf{Disease}  & \textbf{conv1}  & \textbf{conv2} & \textbf{conv3} & \textbf{conv4} & \textbf{conv5} & \textbf{fc6}  & \textbf{CNN} \\ \hline
    Ground Glass & {0.979} & 0.978 & 0.984 & 0.985 & 0.984 & 0.970 & \textbf{0.990}   \\ 
    Reticular & {0.951 }& 0.953 & 0.955 & 0.957 & 0.950 & 0.900 & \textbf{0.964}   \\ 
    Honeycomb & {0.765} & 0.770 & 0.780 & 0.744 & 0.753 & 0.743 & \textbf{0.809}  \\ 
    Emphysema & {0.985} & 0.990 & 0.987 & 0.989 & 0.988 & 0.985 & \textbf{0.995} \\ \hline
    \end{tabular}
\end{table}

\begin{table}
 \caption{F-scores between different layers. Both CNN and multi-variant linear regression regress to ILD pixel numbers.} \label{table:F_pixels_pixels}
\centering
    \setlength{\tabcolsep}{5pt}
    \begin{tabular}{  c | c  c  c  c  c  c  c }
    \hline
    & \multicolumn{6}{c}{\textbf{F-score}} \\ \hline
    \textbf{Disease}  & \textbf{conv1}  & \textbf{conv2} & \textbf{conv3} & \textbf{conv4} & \textbf{conv5} & \textbf{fc6}  & \textbf{CNN} \\ \hline
    Ground Glass & {0.871} & 0.875 & 0.883 & 0.885 & 0.889 & 0.868 & \textbf{0.908}   \\ 
    Reticular & {0.699 }& 0.711 & 0.709 & 0.714 & 0.698 & 0.629 & \textbf{0.719}   \\ 
    Honeycomb & {0.349} & 0.384 & 0.386 & 0.356 & 0.384 & 0.344 & \textbf{0.409}  \\ 
    Emphysema & {0.879} & 0.895 & 0.887 & 0.897 & 0.893 & 0.877 & \textbf{0.915} \\
    Overall & {0.713} & 0.750 & 0.750 & 0.735 & 0.754 & 0.716 & \textbf{0.777} \\ \hline
    \end{tabular}
\end{table}

When evaluated using a smaller ILD dataset, the same as the one used in~\cite{gao2016Multi,depeursinge2012building} of 18k CT slices, FV order-less encoding is effective as demonstrated in Table~\ref{table:AUC_pixels_pixels_workshop}. The unordered pooling operating on the first CNN convolutional layer \emph{conv1} produces the overall best quantitative results, especially for honeycomb. Despite residing in the first layer, the filters and activations on \emph{conv1} are still the integrated parts of a deep network since they are learned through back-propagation from deeper layers. From Table ~\ref{table:AUC_pixels_pixels_workshop}, FV encoding with deeply-learned \emph{conv1} filter activations produces the best ILD classification against FV encoding on other layers and without FV. Nevertheless, for the much larger dataset of 240k CT images under 5-fold CV, the computational complexity of FV encoding becomes the performance bottleneck. It could take an undesirably long time and huge memory requirement to calculate the FV Gaussian components and perform the feature encoding. In our experiments, we randomly select a smaller subset of deep activation features ($\sim$1/3) to calculate FV encoding, which may limit the FV encoding performance. In this setting of sufficiently large amount of data, CNN models without FV encoding perform better.

\begin{table}
\caption{Comparing the AUC values between different layers using a smaller dataset of 18k slices. Both CNN and multi-variant linear regression regress to ILD pixel numbers} \label{table:AUC_pixels_pixels_workshop}
\centering
    \setlength{\tabcolsep}{5pt}
    \begin{tabular}{  c | c  c  c  c  c  c  c }
    \hline
    & \multicolumn{6}{c}{\textbf{AUC}} \\ \hline
    \textbf{Disease}  & \textbf{conv1}  & \textbf{conv2} & \textbf{conv3} & \textbf{conv4} & \textbf{conv5} & \textbf{fc6}  & \textbf{CNN} \\ \hline
    Ground Glass & \textbf{0.984} & 0.955 & 0.953 & 0.948 & 0.948 & 0.930 & 0.943   \\ 
    Reticular & \textbf{0.976 }& 0.958 & 0.954 & 0.951 & 0.950 & 0.939 & 0.917   \\ 
    Honeycomb & \textbf{0.898} & 0.826 & 0.828 & 0.823 & 0.806 & 0.773 & 0.698  \\ 
    Emphysema & \textbf{0.988} & 0.975 & 0.967 & 0.966 & 0.967 & 0.985 & 0.988 \\ \hline
    \end{tabular}
\end{table}


\subsection{Class Balancing}

Even though data is not highly unbalanced, as shown in Table~\ref{table:stats}, using the strategy to balance the classes helps promote the performance considerably. With all the other settings fixed, the integration of the class balancing factors $\beta$ into the CNN loss function (Eq. \ref{eq:classification_balancing}) improves the ILD classification performances on every disease class. The overall F-score also increases by $\sim 5\%$, as shown in Table~\ref{table:F-scores-4k-balancing}. We observe similar a performance boost when Eq. \ref{eq:regression_balancing} is employed.

\begin{table*}
\caption{F-scores of multi-label classification, $T=4000$ pixels, comparing the results with and without class balancing.} \label{table:F-scores-4k-balancing}
\centering
    \setlength{\tabcolsep}{5pt}
    \begin{tabular}{  c | c  c  c  c  c }
    \hline
    & \multicolumn{4}{c}{\textbf{F-score}} \\ \hline
    \textbf{Disease}  & \textbf{Ground Glass}  & \textbf{Reticular} & \textbf{Honeycomb} & \textbf{Emphysema} & \textbf{Overall}\\ \hline
    w/o balancing & {0.8507} & {0.7397} & {0.4308} & 0.8895  & 0.7146  \\    \hline
    w balancing & \textbf{0.8825} &\textbf{ 0.7667} & \textbf{0.5068} & \textbf{0.9361 } & \textbf{0.7656 }\\ \hline
    \end{tabular}
\end{table*}

\subsection{Patch-based Classification Baseline and Processing Time}

The image patch-based classification was the state-of-the-art ILD recognition paradigm. To set up a baseline, we have implemented a standard patch-based algorithm for comparison. The patch-based ILD detection is evaluated on user-defined $32 \times 32$ pixel image patches and predicts a single ILD label for each patch. The image patch-based ILD classification is an easier problem compared to holistic slice-based recognition~\cite{gao2015holistic}. However, it is not suitable to efficiently predict multiple ILD patterns simultaneously that co-occur on a single CT image. As shown in Table~\ref{table:patchtime}, the running time is the heaviest burden of the patch-based methods. We adopt the most commonly used sliding window method to detect all the ILDs within an image. Labels are predicted for image patches sampled at a spatial interval of 10 pixels. The running time for each slice ranges from 4.11 to 32.15 seconds, with a mean of 22.64 seconds, depending on the actual area of lung region on that slice. On the other hand, our proposed holistic method takes less than 0.01 seconds to process a CT slice. It finishes evaluating 50k test slices in $<8$ minutes.

Furthermore, most patch-based methods require the pixel-level lung segmentation as preprocessing. Although the healthy lung segmentation problem is relatively easy to solve, pathological lung segmentation remains an obstacle. The benefit of patch-based classification is that it can explicitly present the location of the ILDs on CT slices. However it is possible to adapt the slice-level CNN models to localize of the underlying diseased regions under a weakly-supervised learning fashion \cite{zhou2016learning}, which provides a way to overcome this drawback with high computational efficiency. We leave this as future work.

\begin{table*}
\caption{Running time (seconds) comparison for the image patch-based classification methods and our proposed holistic approach.} \label{table:patchtime}
\centering
    \setlength{\tabcolsep}{5pt}
    \begin{tabular}{  c | c | c | c | c   }
    \hline

    & \textbf{Patch-based Min}  & \textbf{Patch-based Max} & \textbf{Patch-based Mean} & \textbf{Holistic Detection}\\ \hline
    Time & {4.11} & {32.15} & {22.64} & 0.01   \\    \hline

    \end{tabular}
\end{table*}

%

\section{Conclusion}
In this paper, we present three multi-label deep CNN classification and regression models to accurately recognize potential multiple ILD co-occurrence on an input lung CT slice, in a holistic manner. In contrast to previous image patch based approaches where manual ILD ROIs are given as prerequisites ~\cite{song2015large,anthimopoulos2016lung,van2016combining}, our method performs the task of multi-label, multi-class ILD detection simultaneously, with no image preprocessing on CT slices. Moreover, we also investigate the effectiveness of exploiting the unordered reformation or pooling from deep CNN convolutional activation features via a FV encoding scheme. The proposed algorithms are validated on a publicly available dataset of 658 patients under five-fold cross-validation, achieving high AUC values and F-scores for detecting four main types of ILDs. Our method can be readily adapted to other CAD problems that face similar large spatial and appearance variations. Future work includes performing cross-dataset transfer learning and incorporating weakly-supervised deep CNN approaches to provide ILD localization information on the slices. Last but not least, the flexible holistic CT slice based deep ILD recognition protocol represents a significant step toward clinically useful automated image analyses.

\section*{Acknowledgments}


This research is supported by the NIH Intramural Research Program, the Center for Infectious Disease Imaging, the Imaging Biomarkers and Computer-Aided Diagnosis Laboratory, the National Institute of Allergy and Infectious Diseases and the NIH Clinical Center. We also thank Nvidia for the donation of a Tesla K40 GPU.

\section*{References}

\bibliography{tmi2016_sp}

\end{document}